\documentclass[runningheads]{llncs}

\PassOptionsToPackage{table,xcdraw}{xcolor}
\usepackage{eccv}

\usepackage{eccvabbrv}

\usepackage{graphicx}
\usepackage{booktabs}
\usepackage{multirow}
\usepackage{amsmath}
\usepackage{bm}
\usepackage{subcaption}
\usepackage{tablefootnote}
\usepackage{threeparttable}
\usepackage{algorithm}
\usepackage{algpseudocode}

\usepackage[accsupp]{axessibility}  
\usepackage{hyperref}
\usepackage{orcidlink}

\DeclareMathOperator*{\argmin}{argmin}

\algnewcommand\algorithmicoutput{\textbf{Output:}}
\algnewcommand\Output{\item[\algorithmicoutput]}

\definecolor{someorange}{rgb}{0.773,0.353,0.067}
\definecolor{someblue}{rgb}{0.27, 0.35, 0.760}
\newcommand{\methodplus}{CGM$^\dagger$}

\begin{document}

\title{Curvature-Guided Mixing for MLLM Adaptation}
\titlerunning{Curvature-Guided Mixing for MLLM Adaptation}

\author{Jinglong Yang\inst{1,2}\textsuperscript{\textdagger}\orcidlink{0000-0002-5616-1861} \and
Jiaxuan He\inst{1}\textsuperscript{\textdagger}\orcidlink{0009-0005-1905-6973} \and
Wenjian Huang\inst{1}\orcidlink{0000-0003-2408-8302} \and
Zhan Zhuang\inst{2}\orcidlink{0000-0003-0215-8728} \and
Jianguo Zhang\inst{1}\textsuperscript{*}\orcidlink{0000-0001-9317-0268}}

\authorrunning{J.~Yang et al.}

\institute{Research Institute of Trustworthy Autonomous Systems and Department of Computer Science and Engineering,\\
Southern University of Science and Technology \and
\mbox{Department of Computer Science, City University of Hong Kong}}

\maketitle
\begingroup
\renewcommand{\thefootnote}{\textdagger}
\footnotetext{These authors contributed equally.}
\renewcommand{\thefootnote}{*}
\footnotetext{Corresponding author.}
\renewcommand{\thefootnote}{}
\footnotetext{Funding: This work is supported by National Natural Science Foundation of China (Grant No. 62276121) and Innovation Team and Talents Cultivation Program of National Administration of Traditional Chinese Medicine, No. ZYYCXTD-D-202403.}
\endgroup

\begin{abstract}
Fine-tuning Multimodal Large Language Models (MLLMs) on specialized tasks often leads to catastrophic forgetting of their general capabilities. Existing model merging methods to combat this are often heuristic or use sub-optimal objectives. We propose Curvature-Guided Mixing (CGM), a theoretically grounded framework that merges pre-trained and fine-tuned models. CGM formulates a joint optimization objective and uses a second-order (Hessian) approximation of the loss landscapes to analytically derive an optimal, closed-form ``soft mixing'' ratio. This ratio intelligently blends parameters based on their relative task-specific curvatures. We also introduce \methodplus{}, a robust ``hard mixing'' variant that performs sparse parameter selection guided by a novel, curvature-aware score. Experiments on LLaVA-1.5 and Qwen-2.5VL across multiple downstream tasks show that CGM and \methodplus{} consistently improve the trade-off between task specialization and general knowledge retention over existing methods. Code is available at \href{https://github.com/zzsyjl/CGM-ECCV-2026}{github.com/zzsyjl/CGM-ECCV-2026}.

\end{abstract}

\section{Introduction}
\label{sec:intro}
Multimodal Large Language Models (MLLMs) have emerged as powerful foundation models capable of joint understanding, generation, and planning across vision and language modalities~\cite{llava,blip2,internvl3,qwen25vl,vlp_cen}.
Their impressive performance stems from pre-training on web-scale datasets, which endows them with a broad, general-purpose knowledge base. While this enables strong zero-shot performance on many tasks, adapting these models to specialized or novel domains requires targeted fine-tuning. This adaptation process, however, often leads to a critical issue: the model's newly acquired skills come at the expense of losing its foundational, pre-trained abilities—a phenomenon known as catastrophic forgetting~\cite{investigating,luo2025empiricalstudy,clap4clip}. The central challenge, therefore, is to develop a methodology that can effectively instill new, task-specific knowledge into an MLLM while safeguarding the vast and robust general intelligence it already possesses.

\begin{figure}[t]
  \centering
  \includegraphics[width=0.9\linewidth]{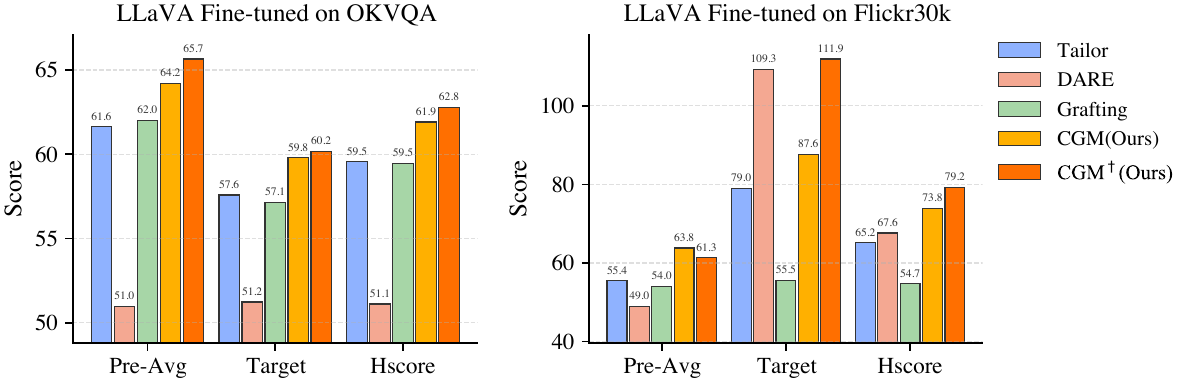}
  \caption{
    Performance comparison of our methods (CGM and \methodplus{}) against baselines for LLaVA fine-tuned on OKVQA. 
    We evaluate general knowledge retention (Pre-Avg: average performance of pre-training tasks), specialization on the new task (Target), and the harmonic mean of both (Hscore) to measure the overall balance.
  }
  \label{fig:okvqa-summary}
\end{figure}

Several recent works have explored this challenge through model merging~\cite{spider,model_tailor} 
and parameter-efficient fine-tuning~\cite{liang2024inflora,sd-lora,wang2025plan,mono_internvl}. 
However, these approaches remain limited in effectively balancing task adaptation and knowledge retention. 
For example, Spider~\cite{spider} merges parameters using heuristic scoring functions without a solid theoretical foundation, while Model Tailor~\cite{model_tailor} employs second-order information~\cite{sparsegpt} but optimizes only for downstream performance.
As a result, these methods tend to overfit to the new task and neglect the preservation of pre-trained general knowledge, underscoring the need for a principled framework that jointly optimizes both objectives.

To overcome these limitations, we introduce \textbf{C}urvature-\textbf{G}uided \textbf{M}ixing (\textbf{CGM}), a theoretically grounded framework for merging pre-trained and fine-tuned models. Our approach begins by formulating a clear objective: to find a new set of weights that simultaneously minimizes the loss with respect to both the fine-tuning and general pre-training tasks. We model the loss landscapes around the fine-tuned and pre-trained optima using a second-order Taylor approximation, which captures the local geometry, or ``curvature'', of each loss surface via its respective Hessian matrix.

\begin{figure}[!t]
\centering
\includegraphics[width=0.6\linewidth]{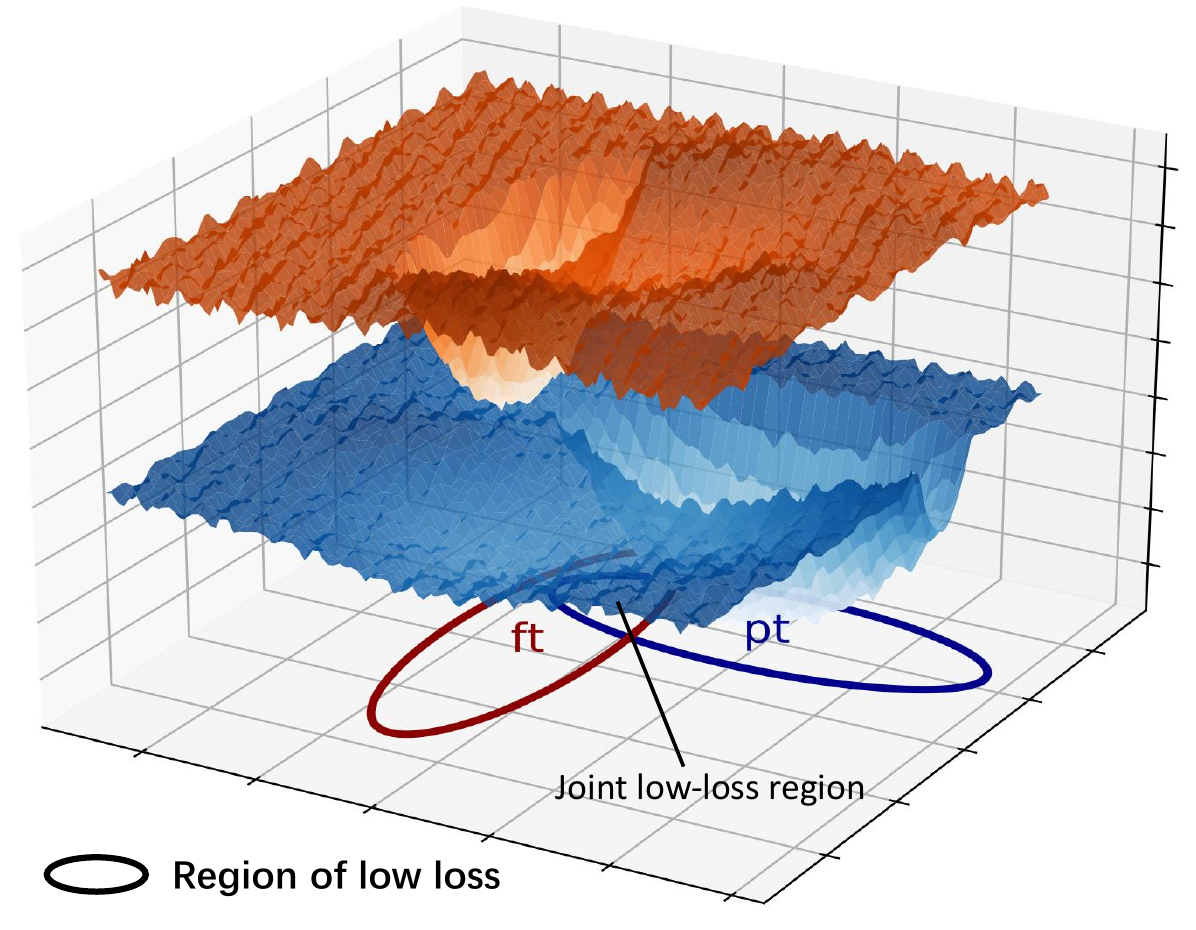}
\caption{
\textbf{A conceptual illustration of the motivation behind CGM.} The fine-tuning loss (orange) and pre-training loss (blue) landscapes exhibit conflicting anisotropic curvatures: the fine-tuning loss is sharp along one axis, whereas the pre-training loss is sharp along the other. CGM leverages this second-order geometry to locate a balanced joint minimum, as detailed in Section~\ref{sec:method}.
}
\vspace{-0.2in}
\label{figs/loss-surfaces.png}
\end{figure}

To illustrate this, consider Figure~\ref{figs/loss-surfaces.png}, which visualizes two hypothetical loss landscapes corresponding to the fine-tuning and pre-training tasks.
A naive interpolation between their optima would likely fall into a region of high loss for both.
Guided by local curvature, our CGM method navigates this anisotropic landscape by identifying that the first parameter direction, critical for the fine-tuning task (due to its high curvature), should be adopted from the fine-tuned model, while the second direction, critical for the pre-training task, should be retained from the pre-trained model.
This enables CGM to effectively integrate the ``skills'' learned during fine-tuning with the foundational knowledge from pre-training, achieving low loss on both tasks and mitigating catastrophic forgetting.

By mathematically formulating and solving this joint optimization problem (as detailed in Section \ref{sec:method}), we derive a closed-form per-parameter mixing ratio. This ``soft mixing'' rule provides a non-heuristic and theoretically grounded solution. It elegantly demonstrates that the optimal mixing ratio for each parameter is directly determined by its relative curvature in the two loss landscapes. 

While soft mixing provides a closed-form solution for the approximated loss, it is fundamentally a dense interpolation that updates every parameter. This approach carries a significant risk: a dense update can be disruptive, potentially destroying existing knowledge. We hypothesize that a sparse update strategy is more robust. The intuition is to preserve the vast majority of the model's parameters and only modify the most critical subset. This approach, which is analogous to imposing an $\mathbf{L_0}$ constraint on the update, ensures a minimal, targeted modification.

To implement this, we propose a ``hard mixing'' strategy, termed \textbf{\methodplus{}}. This method reframes the problem as a sparse selection task. Our method starts from the fine-tuned model—which has acquired the new skill but suffered from forgetting—and treats the pre-trained parameters as a ``knowledge reservoir'' to be sparsely re-integrated. For each parameter, a discrete decision is made on whether to revert to the pre-trained value. We derive a simple, curvature-aware score to rank parameters in the learnable layers. By selecting only the top-$K\%$ of parameters (where $K$ denotes the sparsity ratio) most critical for general knowledge and least critical for the new task, \methodplus{} performs a sparse, targeted reversion that effectively preserves foundational knowledge while maintaining the newly acquired skill and avoiding the pitfalls of dense interpolation.



\begin{figure*}[t]
  \centering
  \includegraphics[width=0.9\textwidth]{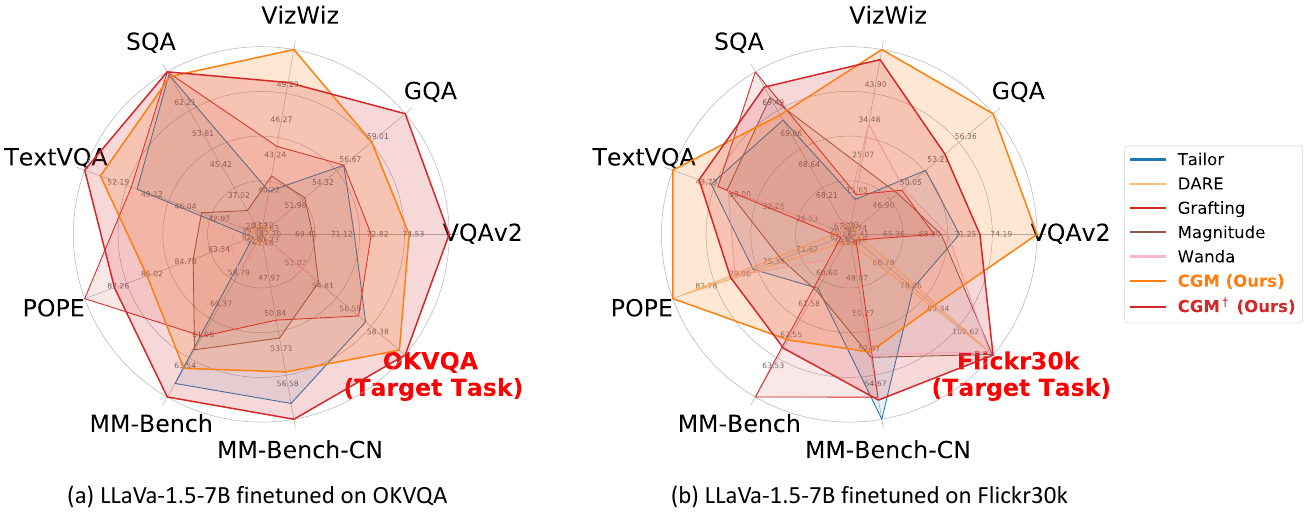}
  \caption{
    Radar plots illustrating the performance 
    trade-off between downstream 
    adaptation
    and general knowledge retention. 
    The ``Target Task'' axis shows performance on the fine-tuned task, 
    while all other axes measure general pre-trained capabilities.
    Our methods, \textbf{CGM} and \textbf{\methodplus{}} demonstrate superior balance by achieving high target-task performance while simultaneously preserving pre-trained knowledge.
  }
  \label{fig:radar_plots}
  \vspace{-5mm}
\end{figure*}

In summary, our main contributions are as follows:

\begin{itemize} 
    \item We propose \textbf{Curvature-Guided Mixing (CGM)}, a novel and theoretically grounded method that optimally merges fine-tuned and pre-trained models by minimizing a joint, curvature-aware loss objective.
    
    \item We introduce \textbf{\methodplus{}}, a variant
    that reformulates the merging task as a sparse parameter selection problem, providing an efficient and effective mechanism for re-injecting pre-trained knowledge to mitigate forgetting. 
    
    \item We conduct extensive experiments across diverse datasets and MLLM backbones, demonstrating that our proposed methods substantially outperform prior approaches in balancing downstream task performance and the preservation of general capabilities. 
\end{itemize}

\section{Related Work}
\label{sec:related_work}

\subsection{Catastrophic Forgetting in MLLMs}
Multimodal Large Language Models (MLLMs)~\cite{llava,blip2,internvl3,qwen25vl} have achieved remarkable success by leveraging large-scale pre-training to acquire broad, general-purpose knowledge. To adapt these models to specialized downstream tasks, fine-tuning is a necessary step~\cite{spider,keep_yourself}. However, this adaptation process often leads to catastrophic forgetting~\cite{investigating,luo2025empiricalstudy}, a well-documented phenomenon where
performance on its original general tasks degrades significantly after learning a new task.

Recent studies have begun to investigate this problem specifically within the MLLM context. Some work focuses on quantifying this effect in continual instruction tuning settings~\cite{coin,continual_llava,climb}, while others observe that anti-forgetting techniques developed for LLMs show limited effectiveness when directly applied to MLLMs~\cite{model_tailor}. 
Our work tackles this core challenge: achieving single-step adaptation to new tasks while rigorously preserving the broad general knowledge obtained during pre-training.

\subsection{Strategies for Knowledge Preservation}
Existing methods to mitigate catastrophic forgetting for MLLM can be broadly categorized into three families:

\noindent\textbf{Regularization-based Methods.}
This classic continual learning approach introduces auxiliary loss terms to penalize significant changes to parameters deemed important for old tasks~\cite{kirkpatrick2017overcoming,zenke2017continual}. However, these methods often require modifications to the fine-tuning loss objective and can be complex to balance with the primary task loss.

\noindent\textbf{Parameter-Efficient Fine-Tuning (PEFT).}
This popular family of methods freezes the vast majority of the pre-trained model and introduces a small set of new, trainable parameters for each task. These can be additive modules like Adapter~\cite{houlsby2019parameter} and LoRA~\cite{hu2022lora}. These approaches are effective at isolating task-specific knowledge and reducing interference (especially in multi-step adaptation~\cite{boosting_moe_clip,pathweave,wu2502,modal-prompt}), and related parameter-efficient editing strategies have also been explored for diffusion models~\cite{chen2025hiding}.

\noindent\textbf{Partial-based Updating and Model Merging.} 
This third category aims to create a merged model by selectively updating a subset of the pre-trained weights. The core challenge lies in determining \textit{which} parameters to update and \textit{how} to merge them. \cite{spider} assesses parameter importance using a heuristic combination of zeroth-order (magnitude) and first-order (gradient) information. Its reliance on a hand-crafted scoring function lacks a rigorous theoretical justification for why their specific formulation is optimal. 
Model Tailor~\cite{model_tailor} leverages the Hessian matrix to identify a sparse ``model patch.'' Inspired by model pruning techniques like SparseGPT~\cite{sparsegpt}, its methodology selects parameters deemed critical for the downstream task. However, this method is designed to minimize loss solely on the fine-tuning task, lacking an explicit objective to balance the preservation of pre-trained knowledge.

In contrast to this prior art, our CGM framework provides a theoretically-grounded solution derived directly from a clear, joint optimization objective. This allows us to derive a non-heuristic, closed-form ``soft mixing'' rule and a ``hard mixing'' score that elegantly uses relative curvature to determine the optimal merge.

\noindent\textbf{Relation to Fisher Merging.}
We also relate our soft mixing to Fisher Merging~\cite{fisher_merging}. Soft mixing resembles Fisher Merging when the curvature is computed using the Fisher Information Matrix (FIM). However, the derivations differ fundamentally: Fisher Merging follows from a Laplace approximation of the posterior, whereas CGM minimizes a joint weighted loss objective. It is theoretically intriguing that distinct assumptions converge to a similar expression. Crucially, CGM is a more general framework: it is not limited to the FIM for curvature estimation, and our further exploration shows, approximating the true Hessian diagonal via Hutchinson trace estimation and Hessian-vector products yields stronger performance.

\vspace{-2mm}
\section{Methodology}
\label{sec:method}

We consider the problem of merging a pre-trained model $\boldsymbol{w^\text{pt}}$ and its fine-tuned variant $\boldsymbol{w^\text{ft}}$ into a unified model $\boldsymbol{w^*}$ that performs well on the target task while preserving general capabilities. In this section, we first formulate this as a joint optimization problem balancing the two loss landscapes, then introduce our method along with its variant.

\begin{algorithm}[t]
\caption{Curvature-Guided Mixing (CGM) Pipeline}
\label{alg:cgm_pipeline}
\begin{algorithmic}[1]
\Require Pre-trained weights $\boldsymbol{w^\text{pt}}$, fine-tuning data $\mathcal{D}_\text{ft}$, calibration set $\mathcal{C}$ (sampling from pre-trained tasks)
\Output Merged weights $\boldsymbol{w^*}$
\State \textbf{Pre-training analysis:} estimate diagonal $\mathbf{H}^\text{pt}$ on $\boldsymbol{w^\text{pt}}$ with $\mathcal{C}$
\State \textbf{Fine-tuning:} SFT on $\mathcal{D}_\text{ft}$ to obtain $\boldsymbol{w^\text{ft}}$, estimate diagonal $\mathbf{H}^\text{ft}$ on $\boldsymbol{w^\text{ft}}$ with $\mathcal{D}_\text{ft}$
\State \textbf{Mixing (choose one):}
\State \quad \textbf{Soft-Mixing (CGM):} compute $\lambda_i$ from Eq.~\ref{eq:soft_mix}, set $w^*_i = w^\text{ft}_i + \lambda_i (w^\text{pt}_i - w^\text{ft}_i)$
\State \quad \textbf{Hard-Mixing (CGM$^\dagger$):} score each parameter with $c_i$ (Eq.~\ref{eq:your_label}); revert the lowest-score $K\%$ of parameters to $w^\text{pt}_i$, keep the rest as $w^\text{ft}_i$
\end{algorithmic}
\end{algorithm}

\subsection{Loss Landscape Approximation}
\label{ssec:loss_approx}
We begin by modeling the local geometry of the loss landscapes around the two optimal weight configurations: $\boldsymbol{w^\text{pt}}$ for general tasks and $\boldsymbol{w^\text{ft}}$ for the specific fine-tuning task. 

The loss functions in the neighborhoods of these optima are approximated using a second-order Taylor expansion. Let $\ell_\text{pt}(\boldsymbol{w})$ and $\ell_\text{ft}(\boldsymbol{w})$ denote the loss functions for the general and fine-tuning tasks, respectively. Their local approximations are given by:
\begin{align}
\ell_{\text{pt}}(\boldsymbol{w}) ={}& \ell_{\text{pt}}(\boldsymbol{w^\text{pt}}) + \nabla \ell_{\text{pt}}(\boldsymbol{w^\text{pt}})^\top (\boldsymbol{w}-\boldsymbol{w^\text{pt}}) \nonumber \\
& \tfrac{1}{2} (\boldsymbol{w}-\boldsymbol{w^\text{pt}})^\top \mathbf{H}^\text{pt} (\boldsymbol{w}-\boldsymbol{w^\text{pt}}) + \mathcal{O}(\Vert\boldsymbol{w}-\boldsymbol{w^\text{pt}}\Vert^3),
\end{align}
\begin{align}
\ell_{\text{ft}}(\boldsymbol{w}) ={}& \ell_{\text{ft}}(\boldsymbol{w^\text{ft}}) + \nabla \ell_{\text{ft}}(\boldsymbol{w^\text{ft}})^\top (\boldsymbol{w}-\boldsymbol{w^\text{ft}}) \nonumber \\
& \tfrac{1}{2} (\boldsymbol{w}-\boldsymbol{w^\text{ft}})^\top \mathbf{H}^\text{ft} (\boldsymbol{w}-\boldsymbol{w^\text{ft}}) + \mathcal{O}(\Vert\boldsymbol{w}-\boldsymbol{w^\text{ft}}\Vert^3),
\end{align}
where $\nabla \ell$ is the first-order gradient vector, $\mathbf{H}$ is the Hessian matrix, and $\mathcal{O}(\cdot)$ represents the higher-order terms.

Our framework relies on two simplifying assumptions.

First, regarding the first-order terms: We assume that $\boldsymbol{w^\text{pt}}$ and $\boldsymbol{w^\text{ft}}$ are local minima 
obtained from training on their respective tasks. By definition, the gradient at any converged local minimum $\boldsymbol{w}_{\text{opt}}$ is zero. 
Therefore, $\nabla \ell_{\text{pt}}(\boldsymbol{w^\text{pt}}) = \mathbf{0}$ and $\nabla \ell_{\text{ft}}(\boldsymbol{w^\text{ft}}) = \mathbf{0}$, allowing the first-order terms to be safely omitted in both Taylor expansions. 


Second, regarding the higher-order terms:
our objective is to find a merged model $\boldsymbol{w}^*$ that balances performance on both the pre-training and fine-tuning tasks. Such a solution must lie within the joint low-loss region—that is, within the neighborhoods of $\boldsymbol{w^\text{pt}}$ and $\boldsymbol{w^\text{ft}}$.
Therefore, a solution $\boldsymbol{w}^*$ within this region satisfies the locality condition of the Taylor expansion, allowing us to omit the higher-order terms $\mathcal{O}(\cdot)$.


Based on these two assumptions and ignoring the constant terms, 
our objective simplifies to minimizing the sum of two quadratic penalties:
\begin{equation}
\ell_{\text{pt}}(\boldsymbol{w}) \propto \tfrac{1}{2} (\boldsymbol{w}-\boldsymbol{w^\text{pt}})^\top \mathbf{H}^\text{pt} (\boldsymbol{w}-\boldsymbol{w^\text{pt}}), \label{eq:pt_approx}
\end{equation}
\begin{equation}
\ell_{\text{ft}}(\boldsymbol{w}) \propto \tfrac{1}{2} (\boldsymbol{w}-\boldsymbol{w^\text{ft}})^\top \mathbf{H}^\text{ft} (\boldsymbol{w}-\boldsymbol{w^\text{ft}}), \label{eq:ft_approx}
\end{equation}
where $\mathbf{H}^\text{pt}$ and $\mathbf{H}^\text{ft}$ denote the Hessian matrices capturing the local curvature of the respective loss surfaces. A larger Hessian eigenvalue corresponds to a sharper curvature, indicating that the loss is more sensitive to perturbations in that parameter direction.


To make the problem computationally tractable, we keep only the diagonal entries (denoted as $h^{\text{pt}}_i$ and $h^{\text{ft}}_i$) of each Hessian and drop all off-diagonal terms.
This simplification, common in curvature-aware optimization and second-order approximation methods~\cite{lecun1990obd,martens2015kfac,kirkpatrick2017overcoming}, reduces the computational and storage cost from $\mathcal{O}(d^2)$ to $\mathcal{O}(d)$ and allows per-parameter decoupling of the objective.

\subsection{CGM: Soft Mixing with Curvature Guidance}
\label{ssec:cgm}
With the quadratic approximations of the loss functions, we define a joint objective to find a new set of weights $\boldsymbol{w}$ that simultaneously minimizes the increase in loss for both tasks:
\begin{equation}
\label{eq:joint_objective}
\argmin_{\boldsymbol{w}} \ell(\boldsymbol{w}) = \tfrac{1}{2} (\boldsymbol{w}-\boldsymbol{w^\text{ft}})^\top \mathbf{H}^\text{ft} (\boldsymbol{w}-\boldsymbol{w}^\text{ft}) + \tfrac{\alpha}{2} (\boldsymbol{w}-\boldsymbol{w}^\text{pt})^\top \mathbf{H}^\text{pt} (\boldsymbol{w}-\boldsymbol{w}^\text{pt}).
\end{equation}
Here, $\alpha > 0$ is a hyperparameter that balances preserving general knowledge (captured by $\mathbf{H}^\text{pt}$) and acquiring new task-specific skills (captured by $\mathbf{H}^\text{ft}$).
This objective seeks a point in the joint low-low region, weighted by their curvature.

We parameterize the merged weights $\boldsymbol{w}$ as a per-parameter linear interpolation starting from the fine-tuned weights. Let $\boldsymbol{\Delta} = \boldsymbol{w^\text{pt}} - \boldsymbol{w^\text{ft}}$ be the ``reversion vector" pointing back to the pre-trained weights. For each parameter $i$, the new weight is $w_i = w^\text{ft}_i + \lambda_i \Delta_i$, where $\lambda_i \in [0,1]$ is the mixing ratio.

Under the diagonal Hessian approximation, the joint objective in Eq.~\ref{eq:joint_objective} decouples into independent per-parameter optimization problems, each expressed as $\min_{\lambda_i} \ell_i(\lambda_i)$:
\begin{equation}
\label{eq:your_label_here}
\min_{\lambda_i} \ell_i(\lambda_i) = \tfrac{1}{2} h^\text{ft}_i ((w^\text{ft}_i + \lambda_i \Delta_i) - w^\text{ft}_i)^2 + \tfrac{\alpha}{2} h^\text{pt}_i ((w^\text{ft}_i + \lambda_i \Delta_i) - w^\text{pt}_i)^2.
\end{equation}
Substituting $w^\text{pt}_i=w^\text{ft}_i + \Delta_i$, this simplifies to:
\begin{equation}
\ell_i(\lambda_i) = \tfrac{1}{2}\Delta_i^2 \big[ h^\text{ft}_i\lambda_i^2 + \alpha h^\text{pt}_i (\lambda_i-1)^2\big].
\end{equation}
This yields a simple quadratic function of $\lambda_i$. By setting the derivative $\partial \ell_i / \partial \lambda_i$ to zero, we obtain the closed-form optimal mixing ratio for each parameter:
\begin{equation}
\label{eq:soft_mix}
\lambda_i^* = \frac{\alpha h^\text{pt}_i}{h^\text{ft}_i + \alpha h^\text{pt}_i}.
\end{equation}
We refer to this result as Curvature-Guided Mixing (CGM),
a theoretically grounded ``soft mixing'' rule.
It states that parameters with higher pre-training curvature ($h^\text{pt}_i$) lead to a larger contribution from the pre-trained model, while those with higher fine-tuning curvature ($h^\text{ft}_i$) retain more from the fine-tuned model.

We are intrigued to observe that if curvature is estimated using the empirical Fisher Information Matrix, the resulting soft-mixing expression closely resembles Fisher Merging~\cite{fisher_merging}. However, the derivations differ fundamentally: Fisher Merging stems from a Laplace approximation of the posterior, whereas CGM minimizes a joint weighted loss objective. It is theoretically interesting that distinct assumptions converge to a similar form. Crucially, CGM is a more general framework and is not tied to any specific Hessian estimation method, and Fisher Merging can be viewed as a degeneralized case of CGM when the Hessian is approximated by the FIM. Our further exploration shows that approximating the true Hessian diagonal via Hutchinson trace estimation and Hessian-vector products yields stronger performance.

\subsection{\methodplus{}: Hard Mixing via Sparse Reversion}
\label{ssec:cgm_plus}

While soft mixing provides a closed-form solution to the approximated objective, it is a dense interpolation, which is often unnecessarily disruptive, as it modifies all parameters, potentially destroying existing knowledge. We argue that a more robust approach is to perform a sparse update: preserve the majority of the model's parameters and only update the most critical subset. To achieve this, we propose a more robust ``hard mixing'' strategy, \methodplus{}, which reframes the problem as a sparse parameter selection task. Instead of blending weights, we make a binary choice for each parameter: either retain the fine-tuned weight $w^\text{ft}_i$ or revert to the pre-trained one $w^\text{pt}_i$.

We introduce a binary mask $\boldsymbol{m} \in \{0,1\}^d$, where $d$ is the number of parameters, and define the merged model as $\boldsymbol{w}(\boldsymbol{m})=\boldsymbol{w^\text{ft}}+\boldsymbol{m}\odot \boldsymbol{\Delta}$, where $\boldsymbol{\Delta} = \boldsymbol{w^\text{pt}} - \boldsymbol{w^\text{ft}}$. Here, $m_i=1$ signifies reverting to the pre-trained weight, and $m_i=0$ signifies keeping the fine-tuned weight. Substituting this into our joint objective (Eq.~\ref{eq:joint_objective}) yields:
\begin{equation}
\ell(\boldsymbol{m}) = \tfrac{1}{2} \sum_i h^\text{ft}_i((w^\text{ft}_i + m_i \Delta_i) - w^\text{ft}_i)^2 + \tfrac{\alpha}{2} \sum_i h^\text{pt}_i((w^\text{ft}_i + m_i \Delta_i) - w^\text{pt}_i)^2.
\end{equation}
\begin{equation}
\ell(\boldsymbol{m}) = \tfrac{1}{2} \sum_i h^\text{ft}_i\Delta_i^2 m_i^2 + \tfrac{\alpha}{2} \sum_i h^\text{pt}_i\Delta_i^2 (m_i - 1)^2.
\end{equation}
Since $m_i \in \{0,1\}$, we have $m_i^2 = m_i$ and $(m_i - 1)^2 = 1 - m_i$. The objective simplifies to:
\begin{equation}
\ell(\boldsymbol{m}) = \tfrac{1}{2} \sum_i h^\text{ft}_i\Delta_i^2 m_i + \tfrac{\alpha}{2} \sum_i h^\text{pt}_i\Delta_i^2 (1 - m_i) .
\end{equation}
Rearranging the terms, we get:
\begin{equation}
\label{eq:your_label}
\ell(\boldsymbol{m}) = \underbrace{\tfrac{\alpha}{2} \sum_i h^\text{pt}_i\Delta_i^2}_{\text{Constant w.r.t. } \boldsymbol{m}} + \sum_i \underbrace{\tfrac{1}{2} (h^\text{ft}_i -\alpha h^\text{pt}_i)\Delta_i^2}_{c_i} m_i.
\end{equation}
Minimizing $\ell(\boldsymbol{m})$ is now equivalent to minimizing $\sum_i c_i m_i$, where $c_i = \tfrac{1}{2}(h^\text{ft}_i - \alpha h^\text{pt}_i)\Delta_i^2$ serves as a ranking score for updating the $i$-th parameter. A parameter with a small $c_i$ (i.e., low $h^\text{ft}_i$ and high $h^\text{pt}_i$) is one that is unimportant for the new task but crucial for the general-purpose pre-trained task, making it a prime candidate for reversion.

To enforce sparsity, we constrain the number of updated parameters to a budget $K$ defined as a sparsity ratio (percentage of parameters updated). Let $d$ be the total number of parameters and $K \in (0,1]$; then $\|\boldsymbol{m}\|_0 = Kd$. The problem becomes:
\begin{equation}
\min_{\boldsymbol{m} \in \{0,1\}^d, \|\boldsymbol{m}\|_0 = Kd} \sum_i c_i m_i.
\end{equation}
This guides us to select the $Kd$ parameters with the smallest $c_i$ values (setting $m_i = 1$) while keeping the rest unchanged ($m_i = 0$). Thus, \methodplus{} performs a sparse and modular update, effectively identifying and applying only the most critical parameter changes from the fine-tuning.

\vspace{-2mm}
\section{Experiments}

\subsection{Experimental Setup}
\label{ssec:experimental_setup}
\noindent\textbf{Architectures and Datasets.} We evaluate our methods on two representative MLLMs: LLaVA-1.5-7B~\cite{llava} and Qwen-2.5VL-3B~\cite{qwen25vl}. For each architecture, we partition the datasets into two categories to separately evaluate generalization capability and downstream adaptability. For LLaVA-1.5-7B~\cite{llava}, we fine-tune on OKVQA~\cite{okvqa} and Flickr30k~\cite{flickr}, representing visual question answering and image captioning, respectively, and evaluate generalization on a standard benchmark suite comprising VQAv2~\cite{1_vqav2}, GQA~\cite{2_gqa}, VizWiz~\cite{3_vizwiz}, SQA~\cite{4_sqa}, TextVQA~\cite{5_textvqa}, POPE~\cite{6_pope}, MM-Bench~\cite{7_mmbench}, and MM-Bench-CN~\cite{8_mmbench-cn}. Similarly, for Qwen-2.5VL-3B~\cite{qwen25vl}, we fine-tune on the Flickr30k~\cite{flickr} and LaTeX-OCR~\cite{latexocr} datasets and assess retained general knowledge on the same benchmark suite, additionally including InfoVQA~\cite{infovqa} and OKVQA~\cite{okvqa} to provide a more comprehensive evaluation of multimodal reasoning and knowledge retention.

\noindent\textbf{Compared Baselines.} We evaluate our method against a diverse set of baselines, including the naive fine-tuning approach and model merging techniques.
\begin{itemize}
\item \emph{Standard Fine-Tuning~\cite{full_ft}:} Full task-specific fine-tuning without anti-forgetting strategies.
\item \emph{Tailor~\cite{model_tailor}:} Merges models by preserving task-critical parameters.
\item \emph{DARE~\cite{dare}:} Uses random parameter selection and rescaling to balance generalization.
\item \emph{Grafting~\cite{grafting}:} Employs skill localization to identify sparse, task-critical weights.
\item \emph{Magnitude:} Heuristically reverts parameters with the smallest absolute changes to pre-trained values.
\item \emph{Wanda~\cite{wanda}:} Reverts parameters based on an importance score of weight magnitude and input activations.
\end{itemize}

\noindent\textbf{Evaluation Metrics.}
 We use (i) the score on the fine-tuning task (reflecting \textit{specialization}) and (ii) the average score across a suite of general pre-training evaluation tasks (reflecting \textit{generalization}). To holistically evaluate effectiveness in mitigating catastrophic forgetting in MLLMs, we adopt two aggregate metrics: \textbf{Average Performance} (Avg) and \textbf{Harmonic Mean Score} (Hscore). 
 Formally, let $S_{\text{Target}}$ denote the performance on the target task, and $S_{\text{Pre-Avg}} = \frac{1}{N}\sum_{i=1}^{N} S_i$ be the average performance over $N$ general pre-training tasks. Then:
\begin{align}
&\text{Avg} = \frac{S_{\text{Target}} + N \cdot S_{\text{Pre-Avg}}}{N+1}, \\
&\text{Hscore} = \frac{2 \cdot S_{\text{Target}} \cdot S_{\text{Pre-Avg}}}{S_{\text{Target}} + S_{\text{Pre-Avg}}}.
\end{align}
Here, the Avg metric equally weights adaptation and generalization, while Hscore penalizes imbalanced performance.

\noindent\textbf{Implementation Details.} We follow the official codebases
to fine-tune LLaVA-1.5-7B~\cite{llava} and Qwen-2.5VL-3B~\cite{qwen25vl}. 
For LLaVA-1.5, we fine-tune the last 12 layers of the language model and the visual projector for 1 epoch with a learning rate of \(1\mathrm{e}{-4}\) and a global batch size of 64.
For Qwen-2.5VL, we fine-tune the last 6 layers and the visual projector for 3 epochs with a learning rate of \(1\mathrm{e}{-5}\) and the same batch size.
We use the empirical Fisher Information Matrix as an efficient estimator of the diagonal Hessian.
All experiments are run on 5 NVIDIA RTX 6000 Ada GPUs (48 GB each).

\noindent\textbf{Computation and Memory Cost.} Obtaining $\boldsymbol{w^{\text{ft}}}$ follows standard supervised fine-tuning. Both CGM (soft mixing) and CGM$^\dagger$ (hard mixing) perform element-wise operations on the selected layers (the last 12 layers for LLaVA and the last 6 layers for Qwen), yielding $\mathcal{O}(d)$ complexity, where $d$ is the number of parameters in those layers. The main additional cost is estimating diagonal Hessians. For $\mathbf{H}^{\text{pt}}$, we use a calibration set sampled from pre-training tasks (8 samples per task in practice) and compute squared gradients at $\boldsymbol{w^{\text{pt}}}$ to obtain an empirical FIM estimate. For $\mathbf{H}^{\text{ft}}$, we accumulate squared gradients over the fine-tuning dataset. Empirically, incorporating FIM estimation during fine-tuning reduces throughput from 3.81 to 3.51 samples/s ($\sim$7.9\% overhead), confirming negligible extra computation. For memory cost, this storage pattern is standard in model-merging methods: we keep only element-wise second-order statistics (the diagonal vectors $\mathbf{h}^{\text{pt}}$ and $\mathbf{h}^{\text{ft}}$). This is substantially more memory-efficient than methods that jointly store first-order, second-order, and other auxiliary statistics.

\subsection{Main Results}
\label{ssec:main_results}
Tables~\ref{tab:llava_okvqa}--\ref{tab:qwen_latex} summarize the main results across multiple datasets and backbones. We evaluate performance using Hscore and Avg, which reflect the trade-off between adaptation to new tasks and retention of general knowledge.

\begin{table*}[!t]
\centering
\caption{Main results on the \textbf{LLaVA-1.5-7B} backbone, fine-tuned on the \textbf{OKVQA} target task.}
\vspace{-0.1in}
\label{tab:llava_okvqa}
\resizebox{\textwidth}{!}{%
\begin{threeparttable}
\begin{tabular}{@{}lcccccccccc|cc@{}}
\toprule
& \multicolumn{9}{c}{\textbf{Pre-trained Tasks}} & \textbf{Target Task} & \multicolumn{2}{c}{\textbf{Overall Metrics}} \\
\cmidrule(r){2-10} \cmidrule(r){11-11} \cmidrule(l){12-13}
Method & VQAv2 & GQA & VizWiz & SQA & TextVQA & POPE & MM-Bench & MM-Bench-CN & \textcolor{someblue}{Pre-Avg} & \textcolor{someorange}{OKVQA} & Hscore & Avg \\
\midrule
Pre-trained Model & 78.5 & 61.9 & 50.0 & 70.4 & 58.2 & 87.3 & 64.3 & 58.3 & \textbf{66.1} & 52.8 & 58.7 & 64.6 \\
Fine-tuned Model & 68.0 & 50.4 & 38.7 & 24.5 & 40.1 & 83.2 & 56.9 & 41.6 & 50.4 & 58.0 & 53.9 & 51.2 \\
\midrule
Tailor & 71.7 & 56.1 & 40.1 & 69.8 & 50.5 & 82.3 & 64.4 & 58.2 & 61.6 & 57.6 & 59.6 & 61.2 \\
DARE & 67.7 & 49.6 & 37.2 & 28.6 & 39.9 & 82.4 & 57.2 & 45.2 & 51.0 & 51.2 & 51.1 & 51.0 \\
Grafting & 72.5 & 56.1 & 44.0 & 70.6 & 51.0 & 88.5 & 62.0 & 51.4 & 62.0 & 57.1 & 59.5 & 61.5 \\
Magnitude & 69.7 & 52.8 & 41.5 & 33.0 & 44.6 & 84.5 & 62.7 & 52.8 & 54.0 & 54.5 & 54.9 & 55.1 \\
Wanda & 68.8 & 51.2 & 39.1 & 30.3 & 41.4 & 83.1 & 58.8 & 45.1 & 52.2 & 53.9 & 53.1 & 52.4 \\
\midrule

\textbf{CGM (Ours)} & 74.3\textsubscript{$\pm$\,0.04} & 58.5\textsubscript{$\pm$\,0.06} & 52.3\textsubscript{$\pm$\,0} \tnote{a} & 69.3\textsubscript{$\pm$\,0.13} & 53.8\textsubscript{$\pm$\,0.35} & 86.2\textsubscript{$\pm$\,0.12} & 63.7\textsubscript{$\pm$\,0.27} & 55.6\textsubscript{$\pm$\,0.40} & 64.2\textsubscript{$\pm$\,0.03} & \underline{59.8\textsubscript{$\pm$\,0.21}} & \underline{61.9\textsubscript{$\pm$\,0.11}} & \underline{63.7\textsubscript{$\pm$\,0.03}} \\
\textbf{\methodplus{} (Ours)} & 76.2\textsubscript{$\pm$\,0.04} & 61.4\textsubscript{$\pm$\,0.07} & 49.5\textsubscript{$\pm$\,0} & 70.6\textsubscript{$\pm$\,0.12} & 55.3\textsubscript{$\pm$\,0.11} & 87.4\textsubscript{$\pm$\,0.20} & 65.1\textsubscript{$\pm$\,0.32} & 59.5\textsubscript{$\pm$\,0.40} & \underline{65.7\textsubscript{$\pm$\,0.05}} & \textbf{60.2\textsubscript{$\pm$\,0.22}} & \textbf{62.8\textsubscript{$\pm$\,0.12}} & \textbf{65.0\textsubscript{$\pm$\,0.07}} \\

\bottomrule
\end{tabular}%
\begin{tablenotes}
      \item[a] The VizWiz server did not respond during our standard-deviation evaluation.
    \end{tablenotes}

\end{threeparttable}
}
\vspace{-3mm}
\end{table*}

\begin{table*}[!t]
\centering
\caption{Main results on the \textbf{LLaVA-1.5-7B} backbone, fine-tuned on the \textbf{Flickr30k} target task.}
\vspace{-0.1in}
\label{tab:llava_flickr}
\resizebox{\textwidth}{!}{%
\begin{tabular}{@{}lcccccccccc|cc@{}}
\toprule
& \multicolumn{9}{c}{\textbf{Pre-trained Tasks}} & \textbf{Target Task} & \multicolumn{2}{c}{\textbf{Overall Metrics}} \\
\cmidrule(r){2-10} \cmidrule(r){11-11} \cmidrule(l){12-13}
Method & VQAv2 & GQA & VizWiz & SQA & TextVQA & POPE & MM-Bench & MM-Bench-CN & \textcolor{someblue}{Pre-Avg} & \textcolor{someorange}{Flickr30k} & Hscore & Avg \\
\midrule
Pre-trained Model & 78.5 & 61.9 & 50.0 & 70.4 & 58.2 & 87.3 & 64.3 & 58.3 & \textbf{66.1} & 11.73 & 19.93 & 60.08 \\
Fine-tuned Model & 67.8 & 46.7 & 22.4 & 67.5 & 37.9 & 76.1 & 59.7 & 50.8 & 53.6 & \textbf{53.57} & 53.59 & 53.6 \\
\midrule
Tailor & 70.8 & 51.7 & 13.2 & 69.3 & 42.9 & 77.6 & 61.1 & 56.9 & 55.4 & 41.85 & 47.69 & 53.92 \\
DARE & 62.4 & 43.7 & 6.2 & 67.8 & 20.8 & 85.3 & 59.6 & 45.9 & 49.0 & 52.48 & 50.66 & 49.36 \\
Grafting & 68.7 & 49.0 & 14.5 & 69.9 & 41.8 & 67.9 & 64.5 & 55.5 & 54.0 & 30.56 & 39.02 & 51.37 \\
Magnitude & 69.3 & 48.6 & 22.1 & 69.6 & 40.2 & 74.5 & 61.2 & 53.0 & 54.8 & \underline{52.78} & 53.77 & 54.57 \\
Wanda & 69.8 & 49.2 & 33.5 & 67.8 & 40.0 & 79.6 & 60.1 & 52.8 & 56.6 & 50.81 & 53.56 & 55.97 \\
\midrule

\textbf{CGM (Ours)} & 77.1\textsubscript{$\pm$\,0.02} & 59.5\textsubscript{$\pm$\,0.09} & 53.3\textsubscript{$\pm$\,0} & 69.3\textsubscript{$\pm$\,0.25} & 49.5\textsubscript{$\pm$\,0.06} & 86.5\textsubscript{$\pm$\,0.26} & 62.7\textsubscript{$\pm$\,0.22} & 52.7\textsubscript{$\pm$\,0.69} & \underline{63.8\textsubscript{$\pm$\,0.08}} & 47.9\textsubscript{$\pm$\,0.74} & \underline{54.71}\textsubscript{$\pm$\,0.17} & \textbf{62.01\textsubscript{$\pm$\,0.03}} \\
\textbf{\methodplus{} (Ours)} & 72.4\textsubscript{$\pm$\,0.02} & 54.0\textsubscript{$\pm$\,0.21} & 50.6\textsubscript{$\pm$\,0} & 69.7\textsubscript{$\pm$\,0.05} & 45.0\textsubscript{$\pm$\,0.30} & 80.1\textsubscript{$\pm$\,0.06} & 63.0\textsubscript{$\pm$\,0.13} & 55.7\textsubscript{$\pm$\,0.29} & 61.3\textsubscript{$\pm$\,0.09} & 51.91\textsubscript{$\pm$\,0.82} & \textbf{56.22\textsubscript{$\pm$\,0.26}} & \underline{60.26\textsubscript{$\pm$\,0.15}} \\

\bottomrule
\end{tabular}%
}
\end{table*}

Across all experiments, existing approaches reveal a clear dilemma: the \emph{Fine-tuned Model} achieves strong performance on the target task but suffers severe degradation on pre-trained tasks, while the \emph{Pre-trained Model} retains general capabilities yet fails to adapt. Methods such as DARE and Wanda partially alleviate this issue but still incur substantial knowledge loss. 
In contrast, our methods \emph{CGM} and \emph{\methodplus{}} achieve the best overall balance. For example, in Table~\ref{tab:llava_okvqa}, \methodplus{} attains the highest Hscore of 62.8 while maintaining nearly full general knowledge (Pre-Avg 65.7). Although the target task scores in Tables~\ref{tab:qwen_flickr} and \ref{tab:qwen_latex} are not the highest, our methods still achieve the best overall metrics, significantly outperforming existing approaches in maintaining both specialization and generalization.

\begin{table*}[!t]
\centering
\caption{Main results on the \textbf{Qwen-2.5VL-3B} backbone, fine-tuned on the \textbf{Flickr30k} target task.}
\vspace{-0.1in}
\label{tab:qwen_flickr}
\resizebox{\textwidth}{!}{%
\begin{tabular}{@{}lcccccccccc|cc@{}}
\toprule
& \multicolumn{9}{c}{\textbf{Pre-trained Tasks}} & \textbf{Target Task} & \multicolumn{2}{c}{\textbf{Overall Metrics}} \\
\cmidrule(r){2-10} \cmidrule(r){11-11} \cmidrule(l){12-13}
Method & VQAv2 & GQA & VizWiz & SQA & TextVQA & POPE & MM-Bench & MM-Bench-CN & \textcolor{someblue}{Pre-Avg} & \textcolor{someorange}{Flickr30k} & Hscore & Avg \\
\midrule
Pre-trained Model & 80.7 & 58.9 & 65.9 & 82.7 & 75.6 & 87.6 & 77.7 & 76.6 & \textbf{75.7} & 30.62 & 43.61 & 70.71 \\
Fine-tuned Model & 72.9 & 46.8 & 57.0 & 79.2 & 66.2 & 83.9 & 75.6 & 63.0 & 68.1 & \textbf{52.15} & 59.05 & 66.3 \\
\midrule
Tailor & 70.3 & 46.4 & 56.1 & 79.7 & 61.2 & 84.6 & 75.6 & 60.7 & 66.8 & 47.97 & 55.85 & 64.73 \\
DARE & 73.4 & 47.8 & 54.2 & 78.2 & 64.0 & 86.6 & 73.7 & 57.0 & 66.9 & 51.04 & 57.89 & 65.11 \\
Grafting & 68.8 & 44.3 & 59.3 & 80.5 & 62.4 & 76.0 & 76.7 & 70.1 & 67.2 & 29.11 & 40.63 & 63.01 \\
Magnitude & 72.0 & 45.8 & 56.4 & 79.9 & 65.7 & 83.2 & 76.2 & 63.7 & 67.8 & 50.85 & 58.13 & 65.95 \\
Wanda & 72.8 & 46.8 & 57.2 & 79.3 & 66.1 & 83.9 & 75.4 & 63.1 & 68.1 & \underline{51.46} & 58.61 & 66.23 \\
\midrule

\textbf{CGM (Ours)} & 79.6\textsubscript{$\pm$\,0.02} & 56.1\textsubscript{$\pm$\,0.14} & 63.5\textsubscript{$\pm$\,0} & 82.0\textsubscript{$\pm$\,0.19} & 73.2\textsubscript{$\pm$\,0.15} & 86.9\textsubscript{$\pm$\,0.10} & 78.1\textsubscript{$\pm$\,0.18} & 76.8\textsubscript{$\pm$\,0.09} & \underline{74.5\textsubscript{$\pm$\,0.08}} & 49.1\textsubscript{$\pm$\,0.26} & \underline{59.2\textsubscript{$\pm$\,0.05}} & \textbf{71.69\textsubscript{$\pm$\,0.04}} \\
\textbf{\methodplus{} (Ours)} & 76.9\textsubscript{$\pm$\,0.02} & 53.0\textsubscript{$\pm$\,0.02} & 61.0\textsubscript{$\pm$\,0} & 82.2\textsubscript{$\pm$\,0.11} & 72.4\textsubscript{$\pm$\,0.71} & 86.5\textsubscript{$\pm$\,0.10} & 78.2\textsubscript{$\pm$\,0.10} & 76.7\textsubscript{$\pm$\,0.20} & 73.3\textsubscript{$\pm$\,0.10} & 50.09\textsubscript{$\pm$\,0.29} & \textbf{59.52\textsubscript{$\pm$\,0.09}} & \underline{70.75\textsubscript{$\pm$\,0.08}} \\

\bottomrule
\end{tabular}%
}
\vspace{-3mm}
\end{table*}

\begin{table*}[!t]
\centering
\caption{Main results on the \textbf{Qwen-2.5VL-3B} backbone, fine-tuned on the \textbf{LaTeX-OCR} target task.}
\vspace{-0.1in}
\label{tab:qwen_latex}
\resizebox{\textwidth}{!}{%
\begin{tabular}{@{}lcccccccccc|cc@{}}
\toprule
& \multicolumn{9}{c}{\textbf{Pre-trained Tasks}} & \textbf{Target Task} & \multicolumn{2}{c}{\textbf{Overall Metrics}} \\
\cmidrule(r){2-10} \cmidrule(r){11-11} \cmidrule(l){12-13}
Method & VQAv2 & GQA & VizWiz & SQA & TextVQA & POPE & InfoVQA & OKVQA & \textcolor{someblue}{Pre-Avg} & \textcolor{someorange}{LaTeX-OCR} & Hscore & Avg \\
\midrule
Pre-trained Model & 80.7 & 58.9 & 65.9 & 82.7 & 75.6 & 87.6 & 61.3 & 56.1 & \textbf{71.1} & 21.1 & 32.6 & 65.6 \\
Fine-tuned Model & 66.8 & 42.9 & 52.7 & 81.8 & 65.9 & 80.8 & 54.8 & 44.1 & 61.2 & 72.7 & 66.5 & 62.5 \\
\midrule
Tailor & 69.7 & 45.5 & 54.4 & 82.3 & 70.4 & 82.3 & 58.2 & 46.7 & 63.7 & 57.0 & 60.2 & 62.9 \\
DARE & 66.3 & 44.1 & 50.9 & 78.5 & 63.3 & 82.9 & 53.1 & 42.7 & 60.2 & 74.8 & 66.7 & 61.8 \\
Grafting & 72.6 & 48.4 & 53.8 & 82.7 & 72.3 & 83.9 & 59.0 & 48.3 & 65.1 & 67.0 & 66.1 & 65.3 \\
Magnitude & 68.5 & 44.3 & 53.5 & 82.4 & 67.6 & 81.4 & 55.7 & 45.3 & 62.4 & \textbf{76.0} & 68.5 & 63.9 \\
Wanda & 67.1 & 43.1 & 52.9 & 82.2 & 66.4 & 81.0 & 55.1 & 44.3 & 61.5 & 74.8 & 67.5 & 63.0 \\
\midrule

\textbf{CGM (Ours)} & 80.9\textsubscript{$\pm$\,0.02} & 58.9\textsubscript{$\pm$\,0.04} & 66.0\textsubscript{$\pm$\,0} & 82.7\textsubscript{$\pm$\,0.22} & 75.8\textsubscript{$\pm$\,0.18} & 87.6\textsubscript{$\pm$\,0.15} & 60.0\textsubscript{$\pm$\,0.46} & 57.1\textsubscript{$\pm$\,0.14} & \textbf{71.1\textsubscript{$\pm$\,0.05}} & 70.9\textsubscript{$\pm$\,1.27} & \underline{71.0\textsubscript{$\pm$\,0.64}} & \textbf{71.1\textsubscript{$\pm$\,0.15}} \\
\textbf{\methodplus{} (Ours)} & 80.4\textsubscript{$\pm$\,0.01} & 57.8\textsubscript{$\pm$\,0.17} & 65.0\textsubscript{$\pm$\,0} & 82.7\textsubscript{$\pm$\,0.16} & 75.5\textsubscript{$\pm$\,0.18} & 86.6\textsubscript{$\pm$\,0.12} & 60.1\textsubscript{$\pm$\,0.16} & 56.6\textsubscript{$\pm$\,0.10} & 70.6\textsubscript{$\pm$\,0.03} & \underline{74.9\textsubscript{$\pm$\,0.09}} & \textbf{72.7\textsubscript{$\pm$\,0.04}} & \underline{71.1\textsubscript{$\pm$\,0.03}} \\
\bottomrule
\end{tabular}%
}
\end{table*}

\subsection{Ablation Studies}
\label{ssec:ablation}

\begin{table*}[!t]
\centering
\caption{Ablation study on the components of the \methodplus{} score $c_i = \tfrac{1}{2}(h^\text{ft}_i - \alpha h^\text{pt}_i)\Delta_i^2$. We compare four variants by selectively omitting components from the score. All variants update the same number of parameters ($K\%$ of the total).}
\vspace{-0.1in}
\label{tab:ablation}
\resizebox{0.85\textwidth}{!}{%
\begin{tabular}{@{}l c cc cc cc@{}}
\toprule
& & \multicolumn{2}{c}{LLaVA - OKVQA} & \multicolumn{2}{c}{LLaVA - Flickr30k}& \multicolumn{2}{c}{Qwen3B - LaTeX-OCR} \\
\cmidrule(lr){3-4} \cmidrule(lr){5-6} \cmidrule(l){7-8}
Variant & Score $c_i \propto$ & Hscore & Avg & Hscore & Avg & Hscore & Avg \\
\midrule
1. Magnitude-only & $\Delta_i^2$ & 54.0 & 54.0 & 53.59 & 53.61 & 67.6 & 62.9 \\
2. Fine-tune+Magnitude & $h^\text{ft}_i\Delta_i^2$ & 53.1 & 52.4 & 53.59 & 53.61 & 67.2 & 62.7 \\
3. Pre-train+Magnitude & $-\alpha h^\text{pt}_i \Delta_i^2$ & \underline{61.7} & \textbf{65.1} & \underline{55.76} & \underline{60.25} & \underline{71.7} & \underline{70.8} \\
\midrule
4. \textbf{\methodplus{} (Full)} & $(h^\text{ft}_i - \alpha h^\text{pt}_i)\Delta_i^2$ & \textbf{62.8} & \underline{65.0} & \textbf{56.22} & \textbf{60.26} & \textbf{72.7} & \textbf{71.1} \\
\bottomrule
\end{tabular}%
}
\end{table*}

We conduct an ablation study to verify the components of the curvature-aware score $c_i = \frac{1}{2}(h^\text{ft}_i - \alpha h^\text{pt}_i)\Delta_i^2$ (Table~\ref{tab:ablation}). Relying solely on magnitude ($c_i \propto \Delta_i^2$) or fine-tuning curvature ($c_i \propto h^\text{ft}_i\Delta_i^2$) yields the lowest Hscores on LLaVA-OKVQA (54.0 and 53.1, respectively) due to severe overfitting and catastrophic forgetting. The pre-training variant ($c_i \propto -\alpha h^\text{pt}_i\Delta_i^2$) preserves general knowledge more effectively, improving the Hscore to 61.7. Ultimately, the full formulation consistently achieves the highest scores across all datasets. By jointly modeling both curvatures, it provides an optimal trade-off between adaptation to new tasks and pre-trained model inertia.

\subsection{Hyperparameter Sensitivity}
\label{ssec:hyperparam_sensitivity}

\begin{figure}[t]
  \centering
  \includegraphics[width=1\linewidth]{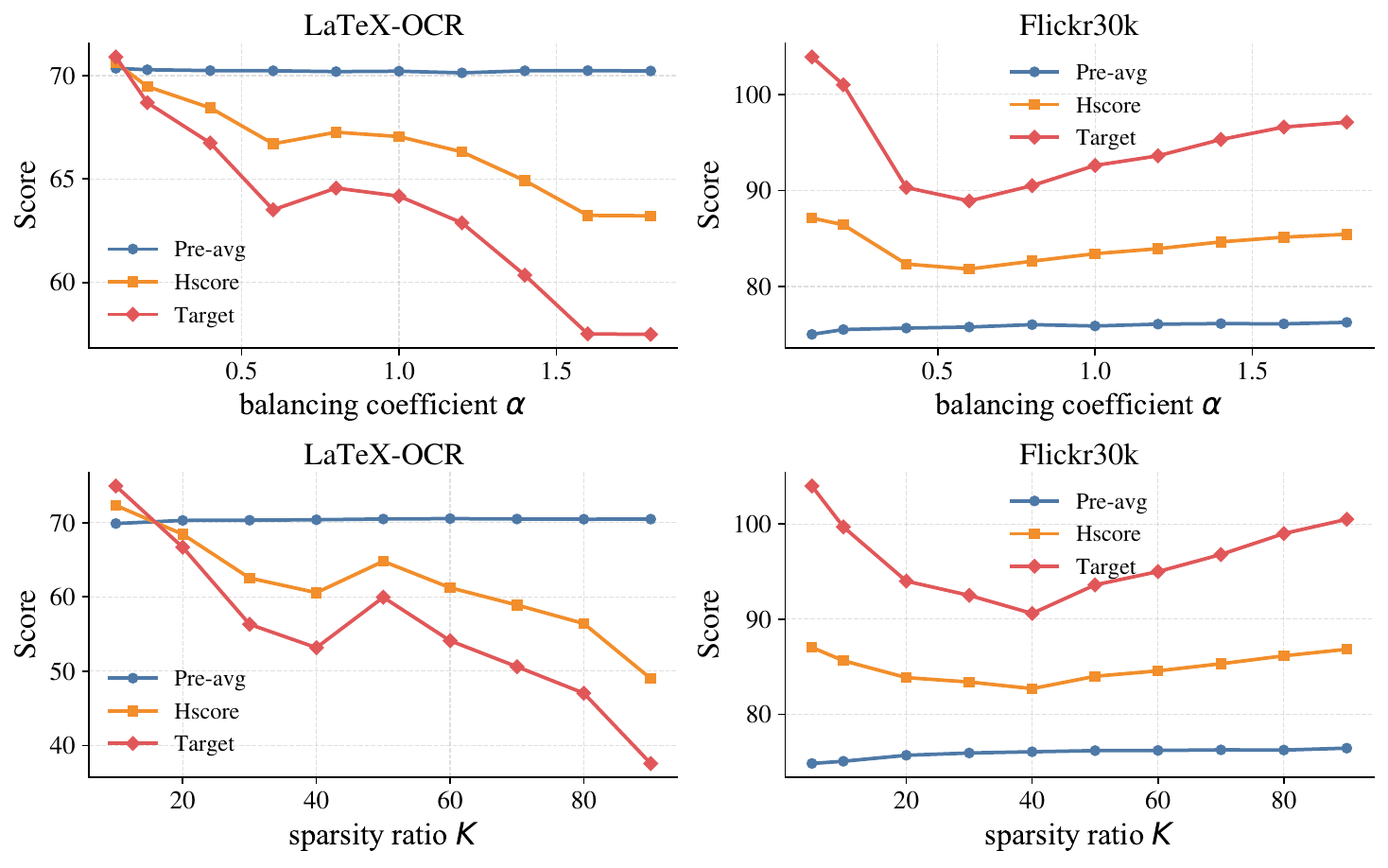}
  \caption{
     Hyperparameter sensitivity analysis on the Qwen3B backbone for the LaTeX-OCR and Flickr30k tasks.
  }
  \label{fig:hyperparam_sensitivity}
\end{figure}

\begin{figure*}[t]
  \centering
  \begin{subfigure}[b]{0.32\textwidth}
    \centering
    \includegraphics[width=\textwidth]{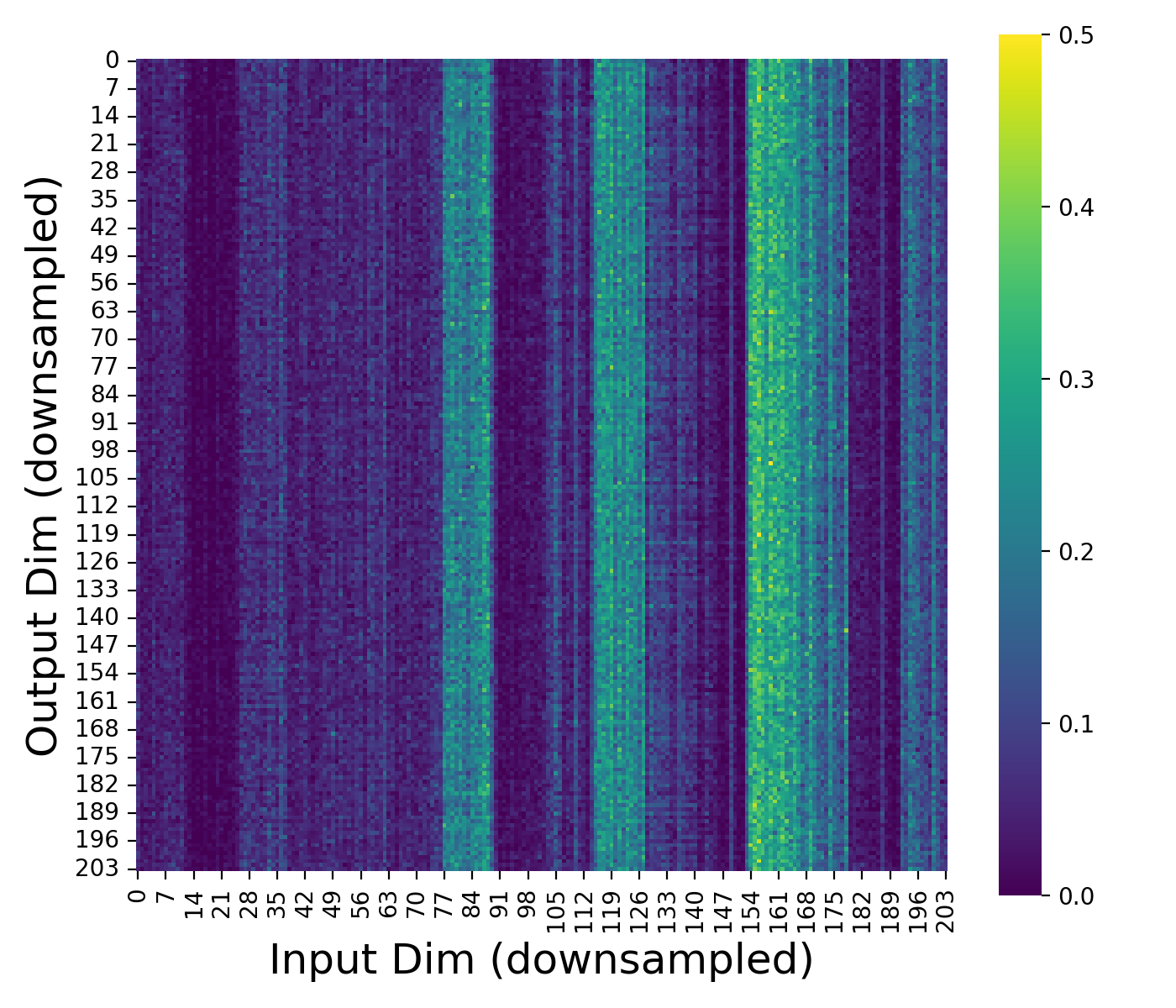}
    \caption{\methodplus{} (Ours)}
    \label{fig:mask_cgm}
  \end{subfigure}
  \hfill
  \begin{subfigure}[b]{0.32\textwidth}
    \centering
    \includegraphics[width=\textwidth]{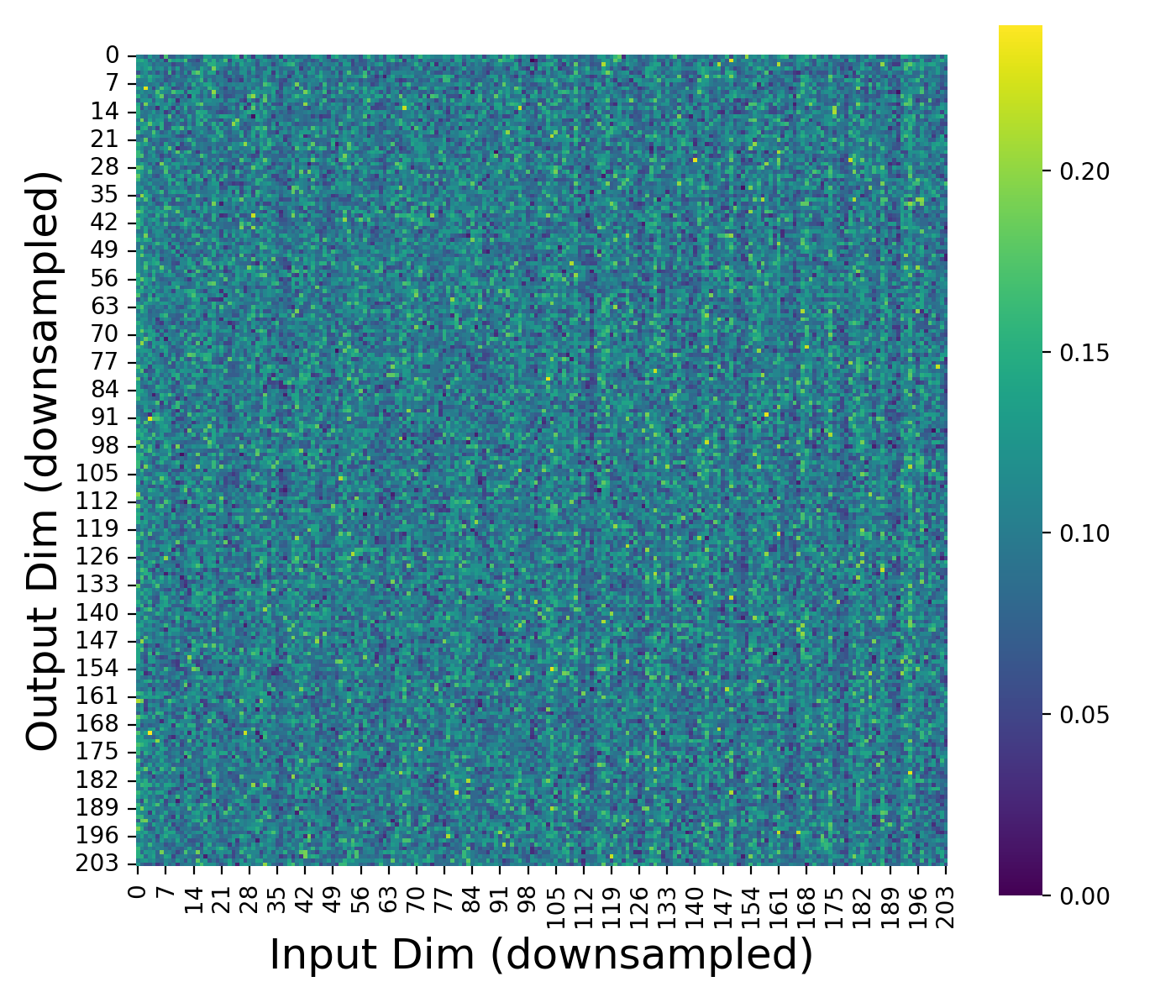}
    \caption{Tailor}
    \label{fig:mask_tailor}
  \end{subfigure}
  \hfill
  \begin{subfigure}[b]{0.32\textwidth}
    \centering
    \includegraphics[width=\textwidth]{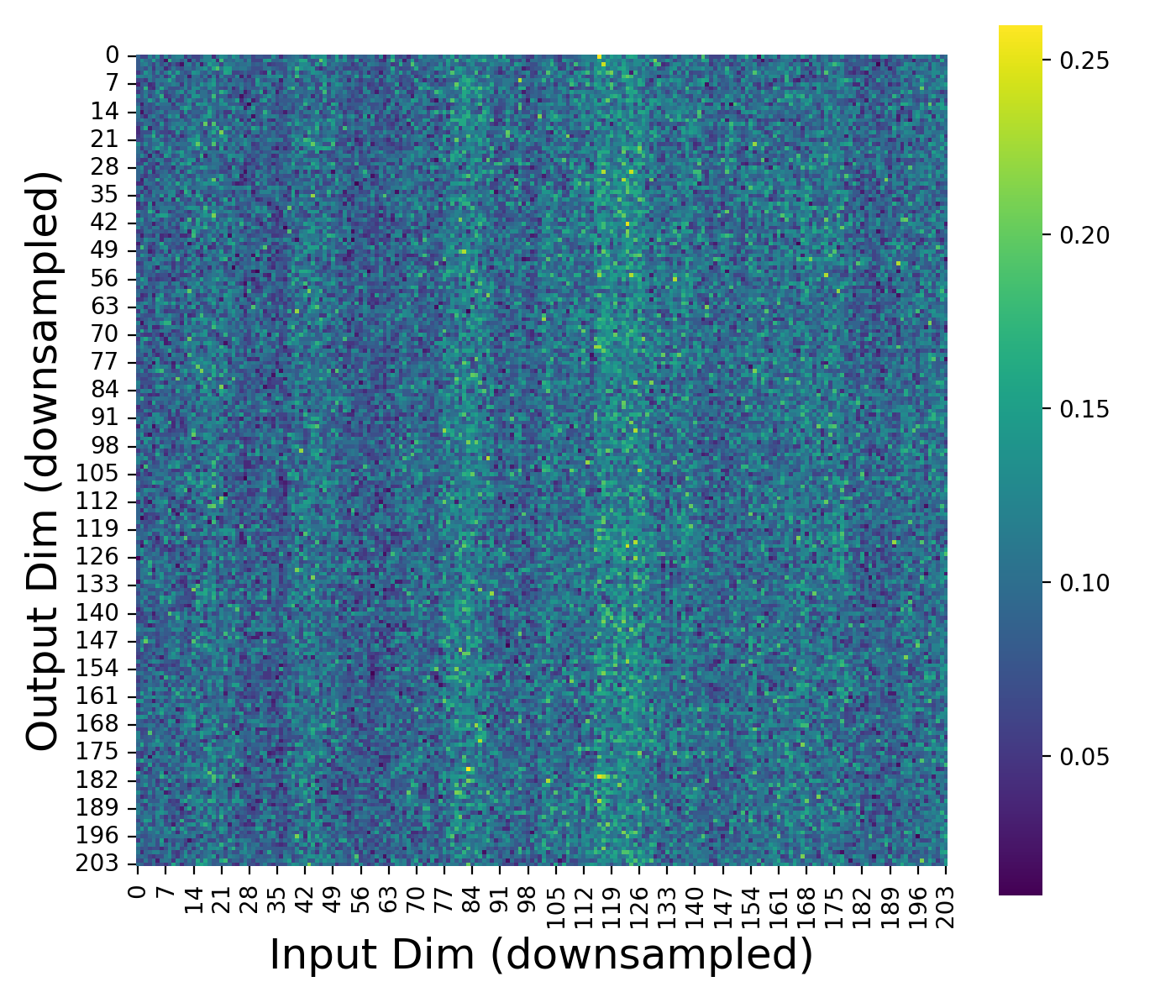}
    \caption{Magnitude}
    \label{fig:mask_mag}
  \end{subfigure}
  \vspace{-0.1in}
  \caption{
    \textbf{Visualization of selection masks (10\% sparsity) for different methods on the} \texttt{self\_attn.o\_proj} \textbf{layer}. The masks are downsampled by patch-averaging for clarity. 
    (a) Our \methodplus{} mask shows a highly structured pattern, selecting by input dimensions (vertical bands). 
    (b) Tailor and (c) Magnitude masks are unstructured and noisy, selecting parameters diffusely across the matrix. This visually confirms that our curvature-guided score identifies coherent structural components for updates.
  }
  \label{fig:mask_vis}
\end{figure*}

\begin{figure*}[t]
  \centering

  \begin{subfigure}[b]{0.32\textwidth}
    \centering
    \includegraphics[width=\linewidth]{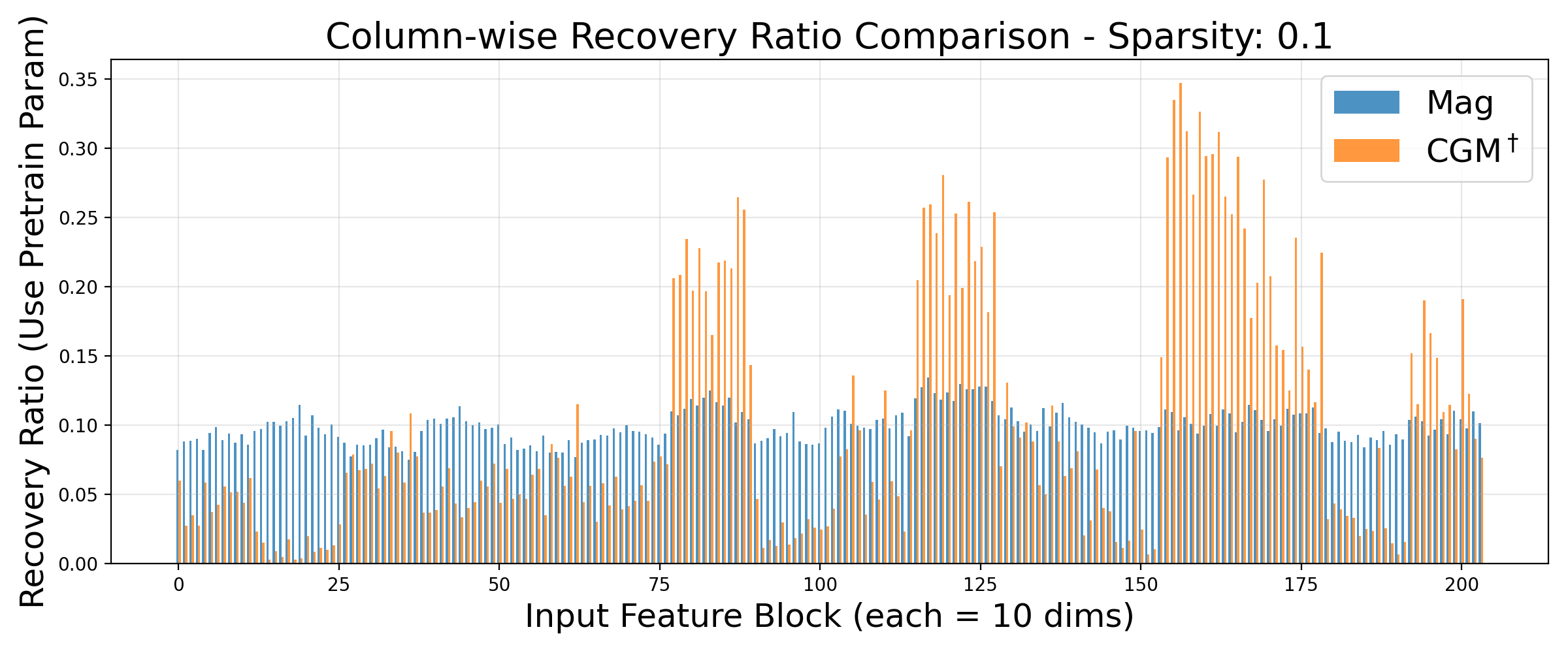}
    \caption{Sparsity = 0.1}
    \label{fig:trend_90_update}
  \end{subfigure}
  \hfill
  \begin{subfigure}[b]{0.32\textwidth}
    \centering
    \includegraphics[width=\linewidth]{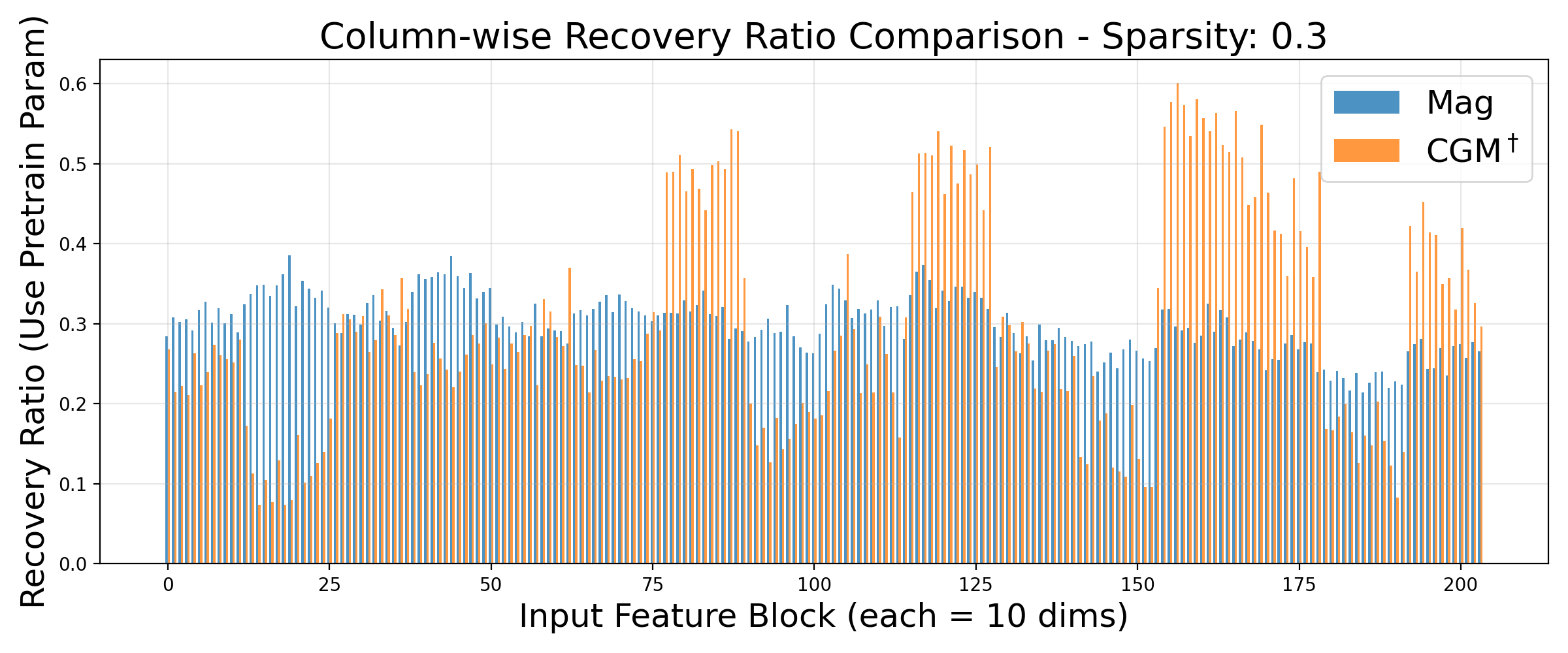}
    \caption{Sparsity = 0.3}
    \label{fig:trend_70_update}
  \end{subfigure}
  \hfill
  \begin{subfigure}[b]{0.32\textwidth}
    \centering
    \includegraphics[width=\linewidth]{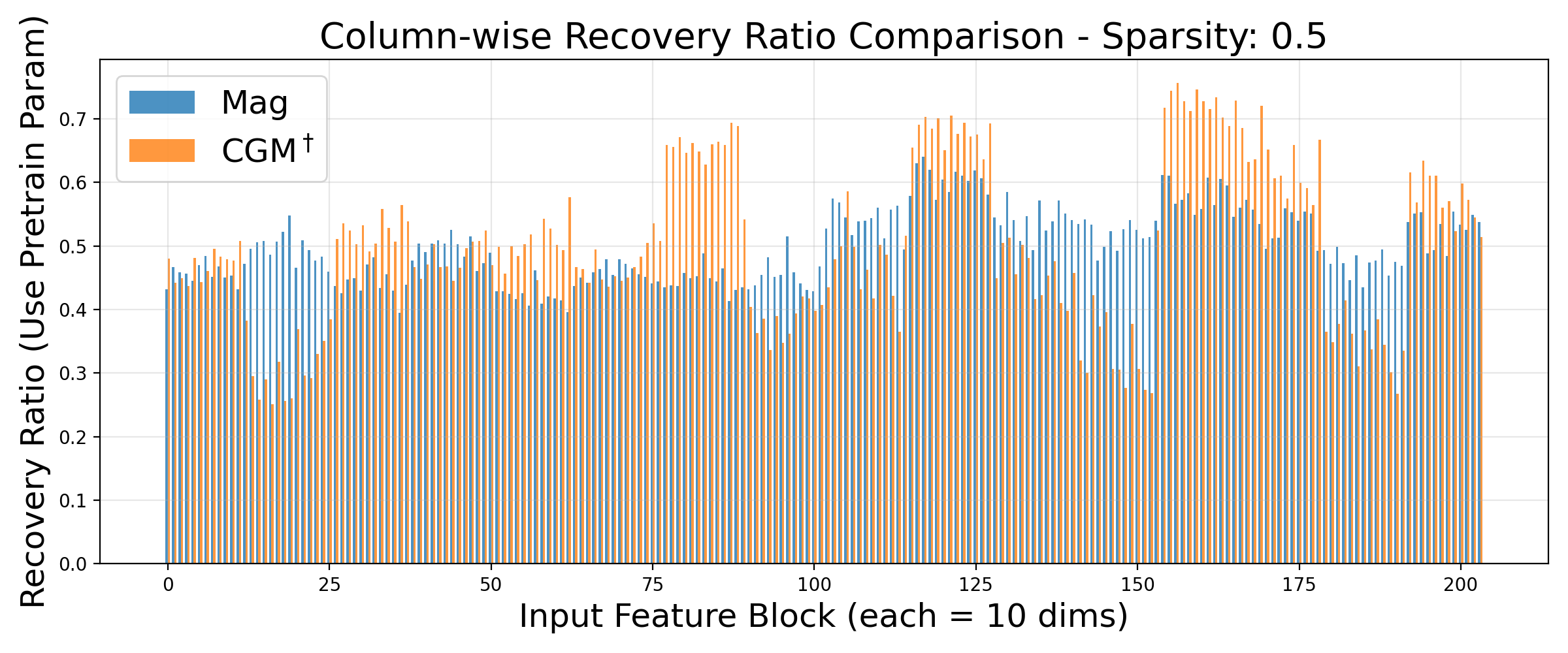}
    \caption{Sparsity = 0.5}
    \label{fig:trend_50_update}
  \end{subfigure}
  \caption{
    Quantitative comparison of column-wise recovery ratios between \methodplus{} (orange) and Magnitude (Mag, blue) at varying update sparsity levels. The Y-axis represents the fraction of pre-trained parameters \textit{kept}.
    Across all sparsity levels, \methodplus{} exhibits a non-uniform, structured selection that consistently targets or protects the same columns, whereas the Magnitude baseline remains uniform and diffuse.
  }
  \label{fig:mask_sparsity_trend}
\end{figure*}


We conduct a sensitivity analysis on our two key hyperparameters, starting with the sparsity ratio $K$, which governs the fraction of pre-trained parameters retained. As shown in Figure \ref{fig:hyperparam_sensitivity}, general knowledge preservation (Pre-Avg score) is remarkably robust, remaining nearly flat even as $K$ varies from 10\% to 90\%. This demonstrates that retaining as few as 10\% of the critically identified pre-trained parameters is sufficient to preserve almost all general knowledge and prevent catastrophic forgetting. Consequently, the primary role of $K$ is to control the degree of downstream specialization rather than managing forgetting, with the optimal Hscore and target performance typically achieved at a 10\% sparsity ratio.

Similarly, the balancing coefficient $\alpha$ modulates the trade-off between target task adaptation and pre-trained knowledge retention. Across varying values of $\alpha$, the Pre-Avg score exhibits minimal variance, confirming that our joint objective inherently shields foundational knowledge. Therefore, $\alpha$ primarily influences target task performance, which in turn dictates the overall Hscore. The best balance is consistently found at smaller values ($\alpha \approx 0.1 \sim 0.15$), which provide sufficient adaptation to the new task while effectively maintaining inherent general capabilities.

\subsection{Analysis of Selection Mask Structure}

To understand the qualitative differences in parameter selection, Figure~\ref{fig:mask_vis} visualizes downsampled masks on the \texttt{self\_attn.o\_proj} layer. Tailor and Magnitude produce diffuse, unstructured noise. In contrast, \methodplus{} exhibits distinct vertical bands, indicating a coherent, feature-level selection focused on specific input dimensions.

We quantify this via the \textit{column-wise recovery ratio} in Figure~\ref{fig:mask_sparsity_trend}. The Magnitude baseline applies updates uniformly across columns at all sparsity levels. Conversely, \methodplus{} consistently targets or protects specific structural column groups (e.g., blocks $\sim$80, $\sim$120, and $\sim$160). This provides strong evidence that our curvature-guided score isolates structured, critically important parameter groups rather than relying on diffuse heuristics.

\section{Conclusion}
In this paper, we introduce Curvature-Guided Mixing (CGM) and its sparse variant, \methodplus{}, as theoretically grounded solutions to catastrophic forgetting in fine-tuned MLLMs. Our framework leverages a joint optimization objective and second-order information from the loss landscapes of both pre-training and fine-tuning tasks. This approach yields an optimal ``soft mixing'' ratio (CGM) and a robust, sparse ``hard mixing'' strategy (\methodplus{}). Extensive experiments on the LLaVA and Qwen-VL models demonstrate that our methods significantly outperform prior art, establishing a new state-of-the-art in balancing task specialization and the preservation of general foundational knowledge. Our results confirm that leveraging loss landscape geometry provides a principled and effective approach to knowledge-preserving model adaptation. 

\clearpage
\bibliographystyle{splncs04}
\bibliography{main}

\end{document}